%% file: main.tex
\documentclass{article}
\usepackage{amssymb}
\usepackage[utf8]{inputenc}

\usepackage[preprint]{corl_2025} 
\usepackage{amsmath}

\usepackage{graphicx}
\usepackage{subcaption}
\usepackage{caption}

\usepackage{multirow}
\usepackage{booktabs}
\usepackage{amsthm}
\usepackage{enumitem}

\usepackage{xcolor}

\usepackage{wrapfig}
\usepackage[export]{adjustbox}

\newtheorem{proposition}{Proposition}

\definecolor{arshblue}{rgb}{0.6, 0.6, 0.8}
\definecolor{robred}{rgb}{1.0, 0.4, 0.4}
\definecolor{nicholsgreen}{rgb}{0.2, 0.8, 0.2}

\definecolor{haojie}{rgb}{0.6, 0.2, 0.8} 

\title{Equivariant Goal Conditioned Contrastive Reinforcement Learning}

\author{
  Arsh Tangri \\ 
  Khoury College of Computer Sciences\\
  Northeastern University, 
  United States\\
  \texttt{tangri.a@northeastern.edu} \\
  \And
  Nichols Crawford Taylor\\
  Khoury College of Computer Sciences\\
  Northeastern University, 
  United States\\
  \texttt{crawfordtaylor.n@northeastern.edu} \\
  \And
  Haojie Huang\\
  Khoury College of Computer Sciences\\
  Northeastern University, 
  United States\\
  \texttt{huang.haoj@northeastern.edu} \\
  \And
  Robert Platt\\
  Khoury College of Computer Sciences\\
  Northeastern University, 
  United States\\
  \texttt{r.platt@northeastern.edu} \\
}

\begin{document}
\maketitle


\begin{abstract}
   Contrastive Reinforcement Learning (CRL) provides a promising framework for extracting useful structured representations from unlabeled interactions. By pulling together state-action pairs and their corresponding future states, while pushing apart negative pairs, CRL enables learning nontrivial policies without manually designed rewards. In this work, we propose Equivariant CRL (ECRL), which further structures the latent space using equivariant constraints. By leveraging inherent symmetries in goal-conditioned manipulation tasks, our method improves both sample efficiency and spatial generalization. 
   Specifically, we formally define Goal-Conditioned Group-Invariant MDPs to characterize rotation-symmetric robotic manipulation tasks, and build on this by introducing a novel rotation-invariant critic representation paired with a rotation-equivariant actor for Contrastive RL. Our approach consistently outperforms strong baselines across a range of simulated tasks in both state-based and image-based settings. Finally, we extend our method to the offline RL setting, demonstrating its effectiveness across multiple tasks.
\end{abstract}



\section{Introduction}

\input{intro}
\vspace{-5pt}
\section{Related Work}
\input{related}
\vspace{-5pt}
\section{Background}
\input{background}

\vspace{-5pt}
\section{Method}
\label{sec:method}
\input{method}

\vspace{-5pt}
\section{Experiments}
\label{sec:experiments}
\input{experiments}

\vspace{-10pt}
\section{Conclusion}
\label{sec:conclusion}
\input{conclusion}




	

\clearpage


\bibliography{nichols_zotero, ref}  
\newpage
\section{Appendix}
\appendix
\input{appendix}

\end{document}

%% file: intro.tex
\vspace{-8pt}
Self-supervised learning has emerged as a pivotal ingredient behind recent scale-driven breakthroughs, where large unlabeled datasets are used to learn powerful representations. However, in the context of reinforcement learning (RL), self-supervision plays a fundamentally different role. Rather than learning from a static dataset, self-supervised RL focuses on learning optimal control-policies through unlabeled sequential interactions with the environment—without relying on manual reward design or human annotation.  Such a learning paradigm can enable scalable robot learning systems that autonomously acquire a broad repertoire of behaviors, generalize across tasks, and adapt to new environments with minimal human intervention \citep{eysenbach_diversity_2018,kim_curiosity-bottleneck_2019,park_metra_2024, zheng_can_2025}. However, achieving this level of autonomy is challenging due to the inherent difficulties of exploration, sparse rewards, and the need for learning robust representations from high-dimensional sensory inputs.

Goal-Conditioned Reinforcement Learning (GCRL) provides a natural framework for this paradigm, as it enables agents to learn to reach states sampled from a goal distribution and can be formulated without requiring externally provided rewards or expert supervision. Recent work \citep{eysenbach_contrastive_2022} has explored the use of contrastive representation learning for GCRL— an approach commonly referred to as Contrastive RL (CRL). This class of methods learns a goal-conditioned Q-function by aligning representations of reachable future states with state-action pairs, while pushing apart unreachable ones. The similarity metric between the state-action and goal embeddings serves as the goal-conditioned Q-value. This enables policy learning without manual reward design and has demonstrated strong performance across a variety of goal-conditioned robotic tasks, positioning CRL as a compelling approach for goal-conditioned RL.

\begin{figure}[t]
  \centering
    \includegraphics[width=0.78\textwidth]{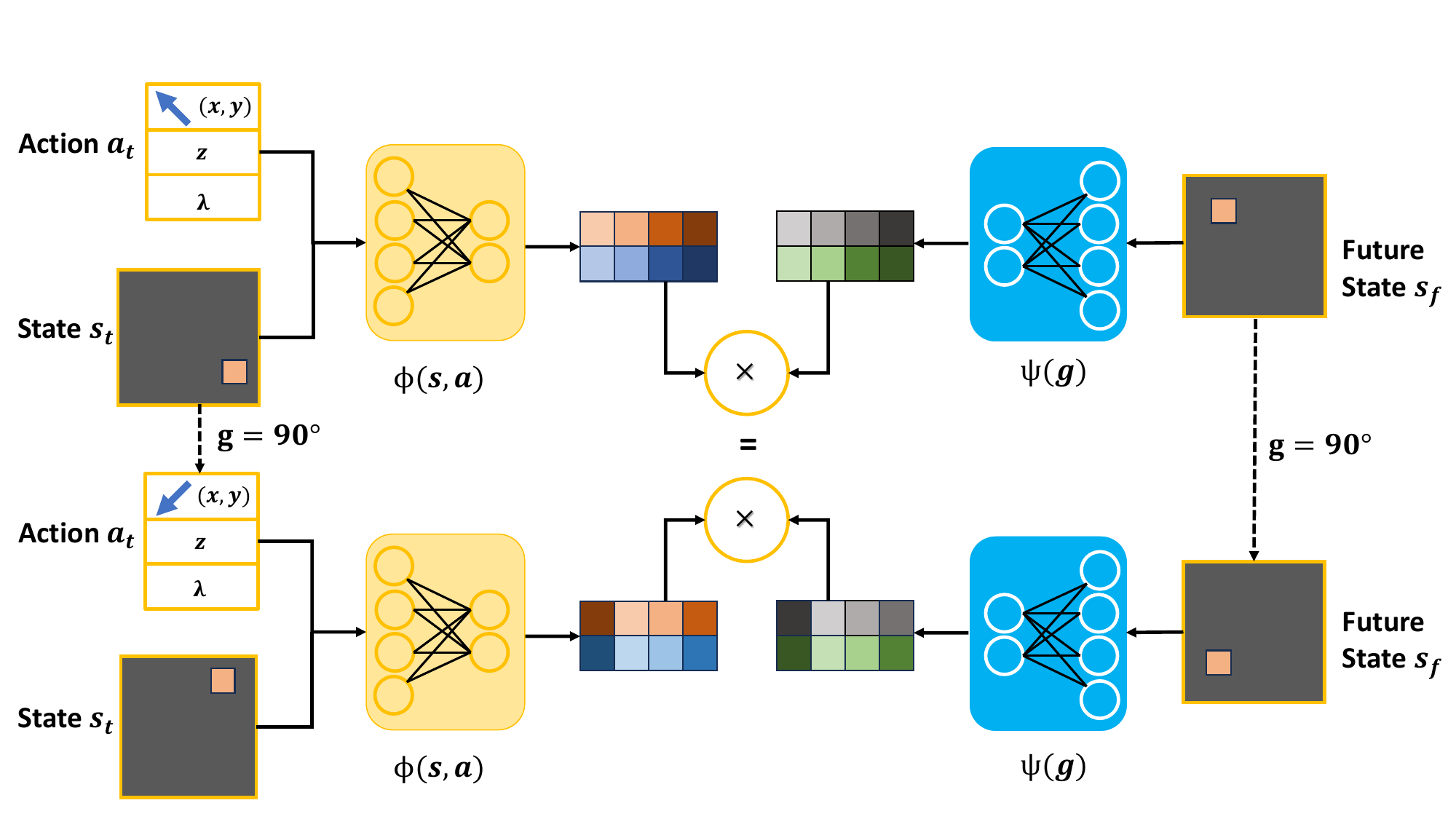}
  \caption{\textbf{Rotation-invariant Critic} The above figure illustrates our rotation-invariant critic, using the $C_4$ group and two channels for simplicity. Each network outputs a stack of two 4-dimensional regular representations of the $C_4$ group. Rotating both the state-action pair $(s, a)$ and the goal $\mathbf{g}$ induces aligned cyclic permutations of each regular-representation vector in their embeddings, but the inner product remains unchanged since corresponding elements are multiplied and summed.}
  \label{fig:Figure-1}
  \vspace{-8pt}
\end{figure}
While Contrastive RL methods have demonstrated improved performance over prior goal-conditioned baselines, they remain sample-inefficient and struggle to learn reliable policies. For example, in our experiments, even after one million environment interactions, CRL fails to reasonably solve manipulation tasks like FetchPush and FetchPickAndPlace. Thus, sample efficiency remains the primary bottleneck in fully unlocking the potential of self-supervised reinforcement learning at scale. To address this limitation, our work incorporates geometric inductive biases—specifically, rotation-equivariance—into Contrastive RL. While leveraging symmetries has been explored in supervised learning~\citep{jia_seil_2023, huang2024leveraging, wang_equivariant_2024} and standard MDPs~\citep{wang_mathrmso2-equivariant_2022}, it has not been directly applied in goal-conditioned reinforcement learning. By integrating equivariance into CRL, we impose group constraints on the learned representations, enhancing generalization and sample efficiency across robotic manipulation tasks.


Our contributions are multifold. First, we analyze the symmetries underlying goal-conditioned manipulation tasks and propose the Goal-Conditioned Group-Invariant MDP by unifying goal-conditioned MDPs and group-invariant MDPs. Second, we introduce a novel approach for constructing a rotation-invariant critic for CRL and combine it with a rotation-equivariant actor to form the Equivariant Contrastive RL algorithm, which explicitly leverages rotational symmetries in robotic manipulation. Third, we demonstrate that our Equivariant Contrastive-RL method improves sample efficiency and generalization across a range of state-based and image-based robotic manipulation tasks, consistently outperforming strong baselines. Finally, we extend Equivariant Contrastive-RL to the offline RL setting and validate its effectiveness across multiple tasks.

%% file: related.tex
\vspace{-8pt}
\textbf{Goal Conditioned RL: }Goal Conditioned Reinforcement Learning~\citep{liu_goal-conditioned_2022, chane-sane_goal-conditioned_2021, steccanella_state_2022} augments standard reinforcement learning with desired goal states. Hindsight Experience Replay~\citep{andrychowicz_hindsight_2017, lin_reinforcement_2019}, or HER, relables past experiences with achieved goals to enable this type of RL. Goal Conditioned Behavior Cloning~\citep{ding_goal-conditioned_2019, lynch_learning_2019, ghosh_learning_2019, srivastava_training_2019} learns goal-reaching policies from expert demonstrations using supervised learning directly.
Contrastive RL~\citep{eysenbach_contrastive_2022, eysenbach_c-learning_2020} reframes goal-conditioned reinforcement learning as a contrastive learning problem, encouraging representations that differentiate achievable future states from random ones. Following this framework, a growing body of work has explored improvements in stability, sample efficiency, and exploration for goal-conditioned contrastive RL \citep{zheng2023stabilizing, zheng2023contrastive, liu_single_2024,
bortkiewicz_accelerating_2024}. Our work builds on this class of methods by integrating rotation-equivariant networks to further improve sample efficiency.

\textbf{Symmetry in Robot Learning: }Symmetry has been leveraged in robot learning to improve both data efficiency and final performance. Prior works~\citep{wang_mathrmso2-equivariant_2022, wang_equivariant_2021, wang_surprising_2022} have leveraged SO(2)-equivariance in policies and invariance in critics through equivariant network architectures~\citep{weiler_general_2019, jenner_steerable_2022}. Equivariant networks have also shown promise for grasp detection, and offline policy learning, leveraging symmetry-aware architectures to enhance performance~\citep{zhu_robot_2023, wang_equivariant_2024, jia_seil_2023, tangri_equivariant_2024, huang2024leveraging}. In this work, We focus on learning equivariant goal-conditioned RL policies without explicit reward supervision.

\textbf{Representation Learning:} High-quality representation learning is a key motivation for CRL. Recent work has also explored representation learning in robotics. For example, ~\citep{ma2022vip, park2024foundation, nair2022r3m} propose methods for learning structured representations that preserve temporal consistency using unlabeled interaction data. Subsequently, ~\citep{jung_yeon_park_learning_2022} introduces symmetry-aware embeddings that transform predictably under group actions. Our work integrates equivariance into Contrastive RL by using rotation-equivariant regular-representation vectors as representations for $(s,a)$ and $\mathbf{g}$. 



%% file: background.tex
\label{sec:background}
\vspace{-8pt}
\textbf{$SO(2)$ and $C_n$ Groups:} Many robotic manipulation tasks exhibit rotational symmetry in the horizontal plane—that is, symmetry with respect to planar rotations around the vertical axis. Formally, the special orthogonal group $SO(2)$ consists of all 2D rotations about the origin, defined as $SO(2) = \{ \text{Rot}_\theta \mid 0 \leq \theta < 2\pi \}$.
Its cyclic subgroup, $C_n$, is a discrete subgroup containing only rotations by angles that are integer multiples of $2\pi/n$, defined as $C_n = \{ \text{Rot}_\theta \mid \theta = \frac{2\pi i}{n},\ i \in {0, 1, \dots, n-1} \}$.

\textbf{Representation of the $C_n$ group:}
A $d$-dimensional representation $\rho: G \rightarrow \mathrm{GL}(d)$ of a group $G$ assigns to each group element $g \in G$ an invertible $d \times d$ matrix $\rho(g)$. Different representations of $C_n$ characterize how various types of signals transform under rotation, including vectors, feature maps, and action parameters. In our setting, $C_n$ acts in three distinct ways relevant to our formulation:

\begin{enumerate}[leftmargin=1em]
    \item \label{item:trivial} \textbf{$\mathbb{R}$ via the trivial representation $\rho_0$:}  
    Let $g \in C_n$ and $x \in \mathbb{R}$. The trivial representation maps every group element to the identity, i.e., $\rho_0(g) x = x$ for all $g \in C_n$.

    \item \label{item:standard} \textbf{$\mathbb{R}^2$ via the standard representation $\rho_1$:}  
    Let $g \in C_n$ and $v \in \mathbb{R}^2$. The standard representation maps $g$ to the 2D rotation matrix corresponding to angle $\theta = \frac{2\pi i}{n}$:
    \begin{equation*}
    \rho_1(g)v = 
    \begin{pmatrix}
    \cos \theta & -\sin \theta \\
    \sin \theta & \cos \theta
    \end{pmatrix} v.
    \end{equation*}

    \item \label{item:regular} \textbf{$\mathbb{R}^N$ via the regular representation $\rho_{\mathrm{reg}}$:}  
    Let $x = \{x_0, x_1, \dots, x_{N-1}\} \in \mathbb{R}^N$ denote a rotation-equivariant feature vector, where each index $i$ corresponds to the group element $\text{Rot}_{2\pi i / N} \in C_N$. The regular representation $\rho_{\mathrm{reg}}$ acts on $x$ by cyclically permuting its elements. Rotations by integer multiples of $\frac{2\pi}{n}$, i.e., $\text{Rot}_{\frac{2\pi i}{n}}$, correspond to $i$ cyclic shifts and are defined as:
    \begin{equation*}
    \rho_{\text{reg}}\left(\text{Rot}_{\frac{2\pi i}{n}}\right)(x_0, x_1, \dots, x_{N-1}) = (x_{N-i}, \dots, x_{N-1}, x_0, \dots, x_{N-i-1}).
    \end{equation*}
\end{enumerate}

\textbf{Group Equivariance and Invariance:} Symmetries in learning problems can be described through the concepts of invariance and equivariance with respect to a group. A function $f$ is said to be invariant under a group $G$ if applying a transformation $g \in G$ to the input $x$ does not change the output, i.e., $f(gx) = f(x)$. In contrast, $f$ is equivariant with respect to $G$ if transforming the input by $g$ results in the output being transformed by the same group element: $f(gx) = g f(x)$.





\textbf{Group-Equivariant Reinforcement Learning}: Equivariant reinforcement learning \citep{wang_mathrmso2-equivariant_2022} leverages the symmetries present in a group-invariant Markov Decision Process (MDP) by structuring the neural networks used to represent the policy and value functions accordingly. An MDP $M = (S,A,R,T,\gamma)$ is said to be a group-invariant ($G-$invariant) MDP if $\forall g \in G$,  $p(s_{t+1} | s_t, a)= p(gs_{t+1} | gs_{t}, ga)$ and $r(s_t, a)= r(gs_t, ga)$.


Importantly, past work \citep{wang_mathrmso2-equivariant_2022} has shown that the optimal solution to an MDP invariant under a group $G$ is also invariant under that group. The optimal Q-function $Q^*$ in a $G$-invariant MDP is itself \textit{group-invariant}, satisfying $Q^*(gs, ga) = Q^*(s, a)$ for all $g \in G$. In contrast, the optimal policy $\pi^*$ is \textit{group-equivariant}, such that $\pi^*(gs) = g\pi^*(s)$.

\textbf{Contrastive RL:} Contrastive RL~\citep{eysenbach_contrastive_2022} uses contrastive learning for Goal-Conditioned Reinforcement Learning (GCRL). A GCRL problem is defined by the goal-conditioned MDP $\langle \mathcal{S}, \mathcal{A}, p(s_0), p(\mathbf{g}), p(s_{t+1} \mid s_t, a_t), r_\mathbf{g}(s_t, a_t) \rangle$, where goals $\mathbf{g} \in \mathbf{G} \subset \mathcal{S}$ each define a reward function $r_\mathbf{g}$ tailored to reaching that goal. We bold $\mathbf{g}$ and $\mathbf{G}$ to distinguish goals from symmetry groups $G$. CRL learns encoders $\phi(s, a)$ and $\psi(\mathbf{g})$, along with a policy $\pi(a \mid s, \mathbf{g})$ that maximizes the likelihood of actions that minimize the distance between the state-action representation and goal representation. A similarity metric defines that distance, $f(s, a, \mathbf{g})$, and $f$ is trained to approximate the Q-function corresponding to a reward defined by the probability of reaching the goal in the next timestep, $r_\mathbf{g}(s, a) := (1-\gamma)p(s_{t+1} = \mathbf{g} \mid s_t, a_t)$ under the current policy.

Prior work has proposed using either the inner product \citep{eysenbach_contrastive_2022}, $f(s, a, \mathbf{g}) := \phi(s, a)^\intercal \psi(\mathbf{g})$, or the $\ell_2$ distance \citep{bortkiewicz_accelerating_2024}, $f(s, a, \mathbf{g}) := -\|\phi(s, a) - \psi(\mathbf{g})\|_2$, as the similarity metric. \citet{eysenbach_contrastive_2022} train the representations using a contrastive binary classification loss \citep{oord_representation_2018, hjelm_learning_2019, ma_noise_2018}, where positive samples $s^+_f$ are taken from the future state distribution given the current state and policy, $p(s_f | s, \pi) $. Negative samples, $s^-_f$, are instead drawn from the future state distribution of random trajectories in the dataset, $p(s_f)$. The loss function is then:
\begin{equation}
\max_{f} \; \mathbb{E}_{(s,a) \sim p(s,a), \; s_f^+ \sim p^\pi(\cdot \mid s,a), \; s_f^- \sim p(s_f)} 
\left[ 
\log \sigma(f(s, a, s_f^+)) + \log (1 - \sigma(f(s, a, s_f^-))) 
\right]
\label{BCE_Contrastive_Loss}
\end{equation}
which minimizes the difference between future state representations and current state and action pairs. The actor is trained similar to other actor critic style algorithms \citep{haarnoja_soft_2018}:
\begin{equation}
\max_{\pi} \mathbb{E}_{p(s)p(g)\pi(a \mid s, g)} \left[ \phi(s, a)^\top \psi(s_f) + \alpha \mathcal{H}\left(\pi(\cdot \mid s, g)\right) \right],
\label{Actor Loss}
\end{equation}

In the offline setting, the policy objective is modified by adding a goal-conditioned behavior cloning loss, similar to \citep{fujimoto2021minimalist}:
\begin{equation}
\max_{\pi(a \mid s, s_g)} \; \mathbb{E}_{\pi(a \mid s, s_g) p(s, a_{\text{orig}}, s_g)} \left[
\lambda \cdot f(s, a, s_f = s_g) + \log \pi(a_{\text{orig}} \mid s, s_g)
\right].
\label{Offline_RL_loss}
\end{equation}


%% file: method.tex
\vspace{-8pt}
The objective of this work is to learn the optimal goal conditioned policy for robotic manipulation tasks that exhibit rotational symmetry in the plane. These tasks can be naturally modeled as goal-conditioned $SO(2)$-invariant Markov Decision Processes (MDPs). Hence, we begin by formally defining the Goal-Conditioned Group-Invariant MDP (GCGI-MDP) and characterizing its properties—showing that the optimal Q-function is group-invariant and the optimal policy is group-equivariant. We then describe planar goal-conditioned robotic manipulation tasks as Goal-Conditioned $SO(2)$-Invariant MDPs, reflecting their inherent rotational symmetry in the plane. To leverage these symmetries, we introduce a rotation-invariant critic and a rotation-equivariant actor within the Contrastive RL framework. We primarily focus on the cyclic group $C_N$, a discrete subgroup of $SO(2)$ consisting of $N$ evenly spaced planar rotations.

\subsection{Goal-Conditioned Group-Invariant MDP}
\vspace{-8pt}
\label{GCGI-MDP}
The ideas of goal-conditioned MDPs and group-invariant MDPs can be naturally extend to a unified setting. A goal-conditioned group-invariant MDP is defined by the tuple:
\[
\mathcal{M}_{G\mathbf{G}} = \langle \mathcal{S}, \mathcal{A}, p(s_0), p(\mathbf{g}), p(s_{t+1} \mid s_t, a_t), r_{\mathbf{g}}(s_t, a_t), G \rangle,
\]
where $\mathbf{g} \in \mathbf{G} \subset \mathcal{S}$ represents a goal and $G$ is the symmetry group. These MDPs must satisfy conditions analogous to standard group-invariant MDPs, ensuring that both the transition dynamics ($\forall g \in G$, $p(s_{t+1} | s_t, a)= p(gs_{t+1} | gs_{t}, ga)$) and the goal-conditioned rewards ($\forall g \in G, r(s_t, a, \mathbf{g})= r(gs_t, ga, g\mathbf{g})$) are invariant under group actions.

\begin{proposition}[Goal-Conditioned Group-Invariant MDP Optimality]
\label{prop1}
Let $\mathcal{M}_{G\mathbf{G}}$ be a group-invariant goal-conditioned MDP. Then its optimal goal-conditioned Q-function is group-invariant,
\begin{equation*}
Q^*(gs, ga, g\mathbf{g}) = Q^*(s, a, \mathbf{g}),
\end{equation*}
and its optimal goal-conditioned policy is group-equivariant for any $g \in G$.
\begin{equation*}
\pi^*(gs, g\mathbf{g}) = g\pi^*(s, \mathbf{g}).
\end{equation*}
\end{proposition}

Proposition~\ref{prop1} formalizes the core principle underlying our method: the optimal goal-conditioned Q-function is rotation-invariant, and the optimal policy is rotation-equivariant under the symmetry group of the environment. It motivates learning in a low-dimensional space defined by the equivalence classes of samples under the group action. Proof can be found in Appendix.

 
\subsection{Goal-Conditioned $SO(2)$-Invariant MDP for Robotic Manipulation}
\vspace{-8pt}
\label{sec:GC-G-invariant MDP}
We first describe the goal-conditioned $SO(2)$-invariant MDP formulation used for our robotic manipulation tasks. In practice, we work with its discrete counterpart, the goal-conditioned $C_N$-invariant MDP, which is defined analogously by restricting the symmetry group to the cyclic subgroup $C_N \subset SO(2)$. Our experiments explore both state-based and image-based settings, covering two complementary input modalities commonly used in robot learning. For the state-based tasks, we assume a factored state and goal space, i.e., $S = S_{\text{inv}} \times S_{\text{equi}} \subseteq \mathbb{R}^k$ and $\mathbf{G} = \mathbf{G}_{\text{inv}} \times \mathbf{G}_{\text{equi}} \subseteq \mathbb{R}^k$. The elements in $S_{\text{inv}}$ and $\mathbf{G}_{\text{inv}}$ are invariant under the action of $g \in C_N$, while the elements in $S_{\text{equi}}$ and $\mathbf{G}_{\text{equi}}$ transform according to the representation $\rho_{\text{equi}} = \rho_1$. For image-based tasks, the state of the environment is represented by a 3-channel RGB image. A group element $g \in C_N$ acts on the image as follows:
\begin{equation}\label{equation:TransformationImage}
(g f_s)(x, y) = \rho_{0}(g)\, f_s(\rho_{1}^{-1}(x, y))
\end{equation}
Here, $\rho_1$ denotes the spatial rotation applied to pixel coordinates, while $\rho_0$ indicates that the channel dimensions remain unchanged under the group action. The MDP is assumed to have a factored action space $A = A_{\text{inv}} \times A_{\text{equi}} \subseteq \mathbb{R}^k$. A group element $g \in C_N$ transforms an action $a = (a_{\text{equi}}, a_{\text{inv}})$, where $a_{\text{equi}} \in A_{\text{equi}}$ and $a_{\text{inv}} \in A_{\text{inv}}$, as: 
\begin{equation}\label{equation:TransformationAction}
g a = (\rho_{\text{equi}}(g)\, a_{\text{equi}},\; a_{\text{inv}})
\end{equation}
An intuitive example of the action of $g \in C_N$ on a factored vector space is the robot action vector $(x, y, z, \theta, w)$ used in our PyBullet tasks \citep{wang_bulletarm_2022}, where $(x, y)$ denotes planar displacement, $z$ is vertical movement, $\theta$ is the gripper rotation angle, and $w$ represents the gripper open/close state. Under the group action, the planar components $(x, y)$ transform according to the standard representation $\rho_1(g)$, while $z$, $\theta$, and $w$ remain unchanged under the trivial representation $\rho_0(g)$, as described in Section~\ref{sec:background}. 
\subsection{Equivariant Contrative RL}
\vspace{-8pt}
Building upon our Goal-Conditioned $SO(2)$-Invariant MDP formulaton for manipulation tasks in Section~\ref{sec:GC-G-invariant MDP}, we propose Equivariant Contrastive RL, which combines a rotation-invariant critic with a rotation-equivariant policy to solve goal-conditioned tasks more efficiently by exploiting the environment's underlying symmetry.

\textbf{Rotation invariant Critic.}
As shown in Figure~\ref{fig:Figure-1}, our rotation-invariant contrastive critic is composed of three components: a state-action encoder $\phi$, a goal encoder $\psi$, and an energy function $f_{\phi,\psi}(s, a, \mathbf{g})$ that measures the similarity between the encoded state-action pair $\phi(s, a)$ and the goal representation $\psi(\mathbf{g})$. In this work, we focus on the inner-product and the $l_2$-distance as similarity measures between the two representations. For the critic to be invariant to $g \in G$, it must satisfy the following condition:
\begin{equation}
f_{\phi,\psi}(s, a, \mathbf{g}) = f_{\phi,\psi}(gs, ga, g\mathbf{g})
\label{eq:inv}
\end{equation}
To satisfy the invariance condition in Equation~\ref{eq:inv}, we represent both the state-action encoder $\phi$ and the goal encoder $\psi$ using $C_N$-equivariant networks. These networks map their inputs onto a stack of $C_N$-equivariant feature representations ($1\times1$ regular representation feature maps, i.e., equivariant vector features): $\phi(s, a) \in \mathbb{R}^{N \cdot K}$ and $\psi(\mathbf{g}) \in \mathbb{R}^{N \cdot K}$, where $N$ is the cardinality of the group $G$, and K is the number of regular-representation vectors. Each regular representation feature vector contains $N$ elements and exhibits a cyclic permutation structure under group actions, as described in Section \ref{sec:background}, which we exploit to achieve rotation-invariant similarity between state-action and goal embeddings. 

This equivariant structure ensures that when the state-action and goal inputs are rotated by the same group element, their corresponding equivariant feature vectors undergo aligned cyclic permutations. As a result, both inner product and $\ell_2$ similarity computations operate on matching components, making the similarity measure invariant to rotations. 
This behavior is illustrated in Figure~\ref{fig:Figure-1}, which shows an invariant critic where each network outputs multiple regular-representation features. Rotating both the state-action pair $(s, a)$ and the goal $\mathbf{g}$ by the same group element induces identical permutations within each regular representation.
However, the similarity metric remains unchanged since corresponding elements are aligned under the group action. This property holds for both inner product and $\ell_2$ distance.

\textbf{Rotation Equivariant Actor.}
The equivariant actor $\pi(a|s, \mathbf{g})$ is also parameterized by a $C_N$-equivariant neural network and outputs a mixed representation type $1 \times 1$ feature map, which represents the action, similar to \citet{wang_mathrmso2-equivariant_2022}. The rotation-equivariance of our network ensures that rotating the input results in a correspondingly rotated output for the equivariant components of the action (e.g., $x$, $y$), while the invariant components (e.g., $z$, $\theta$, gripper state) remain unchanged.

Finally, we train the critic networks using the Binary NCE loss (Eqn.~\ref{BCE_Contrastive_Loss}), and the actor by maximizing the Q-values (Eqn.~\ref{Actor Loss}).




%% file: experiments.tex
\vspace{-8pt}
We evaluate Equivariant Contrastive RL (ECRL) on both state-based and image-based goal-conditioned manipulation tasks across a diverse set of simulated robotics benchmarks.  Performance is measured by rolling out policies for 50 randomly sampled goals during training for each task. In particular, we assess two variants of ECRL that differ in the similarity metric used in the critic: one based on inner product, and the other on $\ell_2$ distance. We select TD3+HER~\citep{andrychowicz_hindsight_2017}, CRL~\citep{eysenbach_contrastive_2022},and a variant of CRL using $\ell_2$ distance~\citep{bortkiewicz_accelerating_2024} as our baselines. We further evaluate ECRL in the Offline-RL setting
and conduct an ablation study to assess the impact of our invariant critic design.
Throughout all experiments, we use the cyclic group $C_8$ as a discrete approximation of $SO(2)$, corresponding to eight evenly spaced planar rotations. Our networks use $C_8$-equivariant MLPs for state-based tasks and $C_8$-equivariant convolutional encoders for image-based tasks, with equivariant layers applied to the group-structured components where appropriate.

\begin{figure}[t]
  \centering
  \begin{subfigure}[b]{0.11\textwidth}
    \includegraphics[width=\linewidth,valign=t]{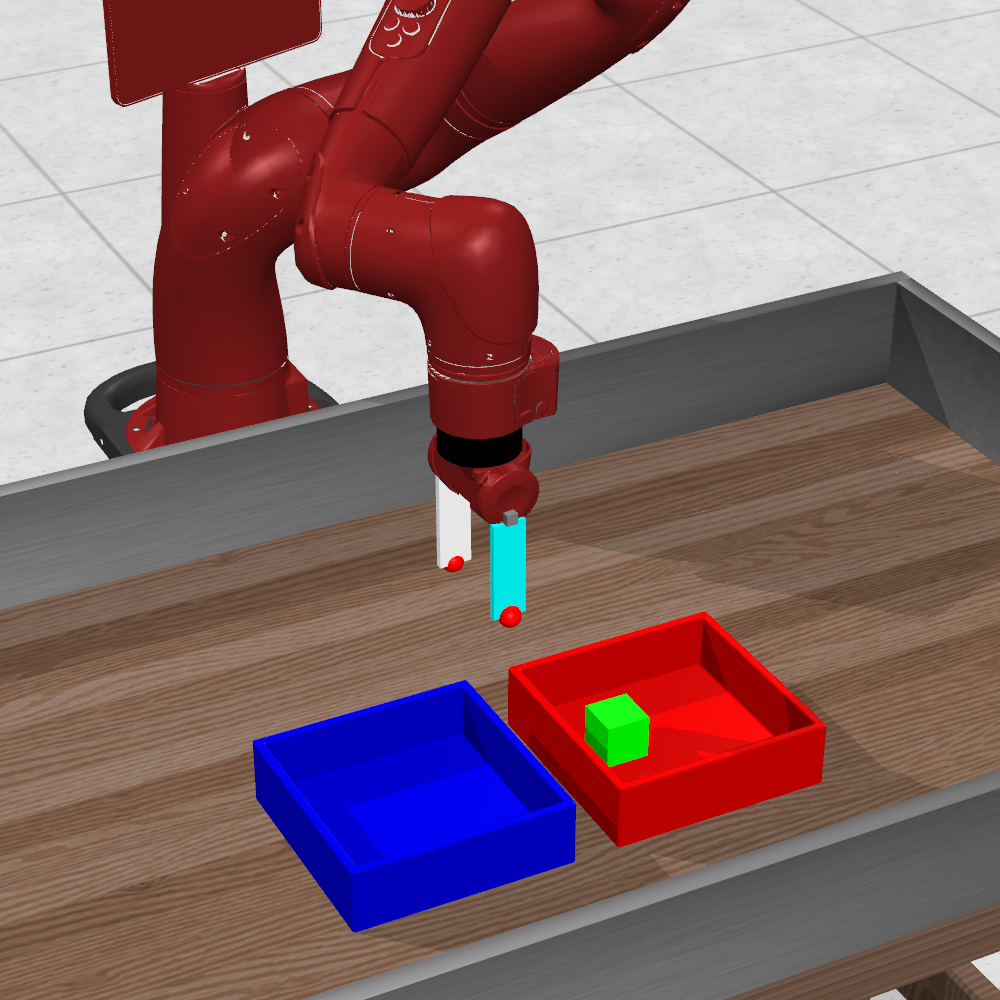}
    \caption{}
    \label{fig:sub1}
  \end{subfigure}
  \begin{subfigure}[b]{0.11\textwidth}
    \includegraphics[width=\linewidth,valign=t]{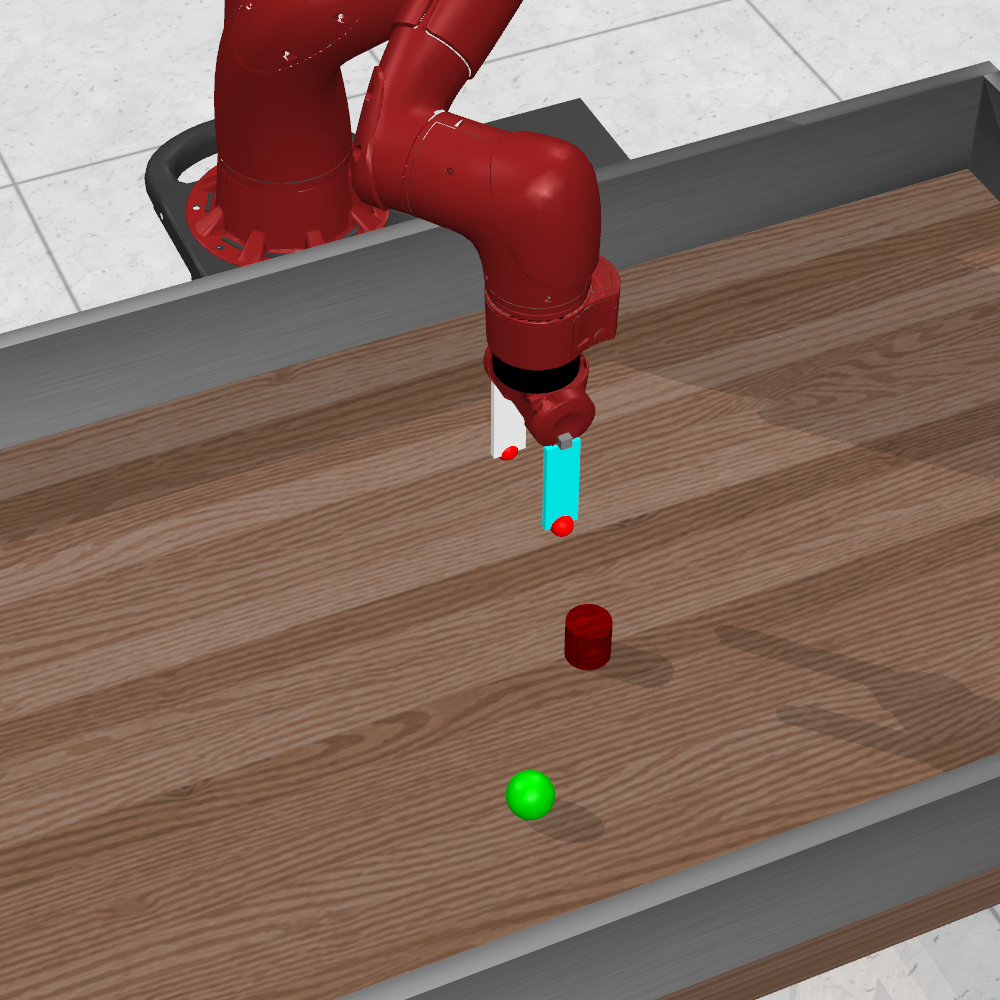}
    \caption{}
    \label{fig:sub2}
  \end{subfigure}
  \begin{subfigure}[b]{0.11\textwidth}
    \includegraphics[width=\linewidth,valign=t]{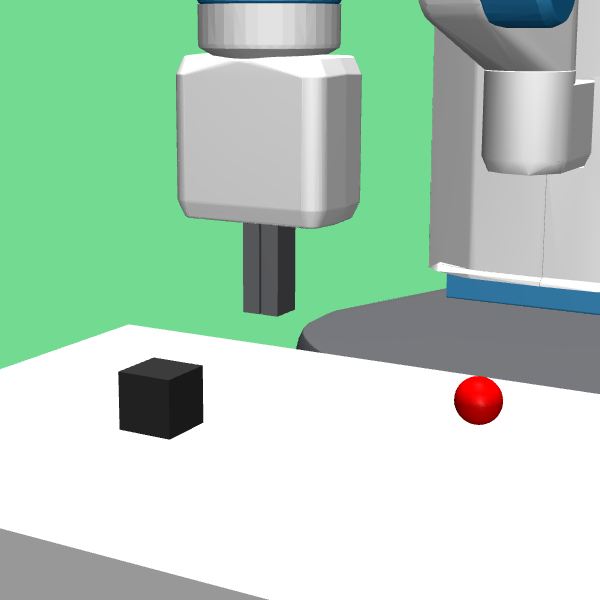}
    \caption{}
    \label{fig:sub3}
  \end{subfigure}
  \begin{subfigure}[b]{0.11\textwidth}
    \includegraphics[width=\linewidth,valign=t]{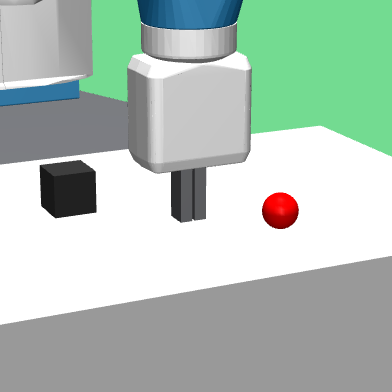}
    \caption{}
    \label{fig:sub4}
  \end{subfigure}
  \begin{subfigure}[b]{0.11\textwidth}
    \includegraphics[width=\linewidth,valign=t]{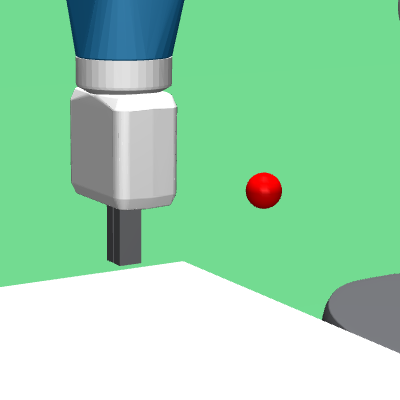}
    \caption{}
    \label{fig:sub5}
  \end{subfigure}
  \begin{subfigure}[b]{0.11\textwidth}
    \includegraphics[width=\linewidth,valign=t]{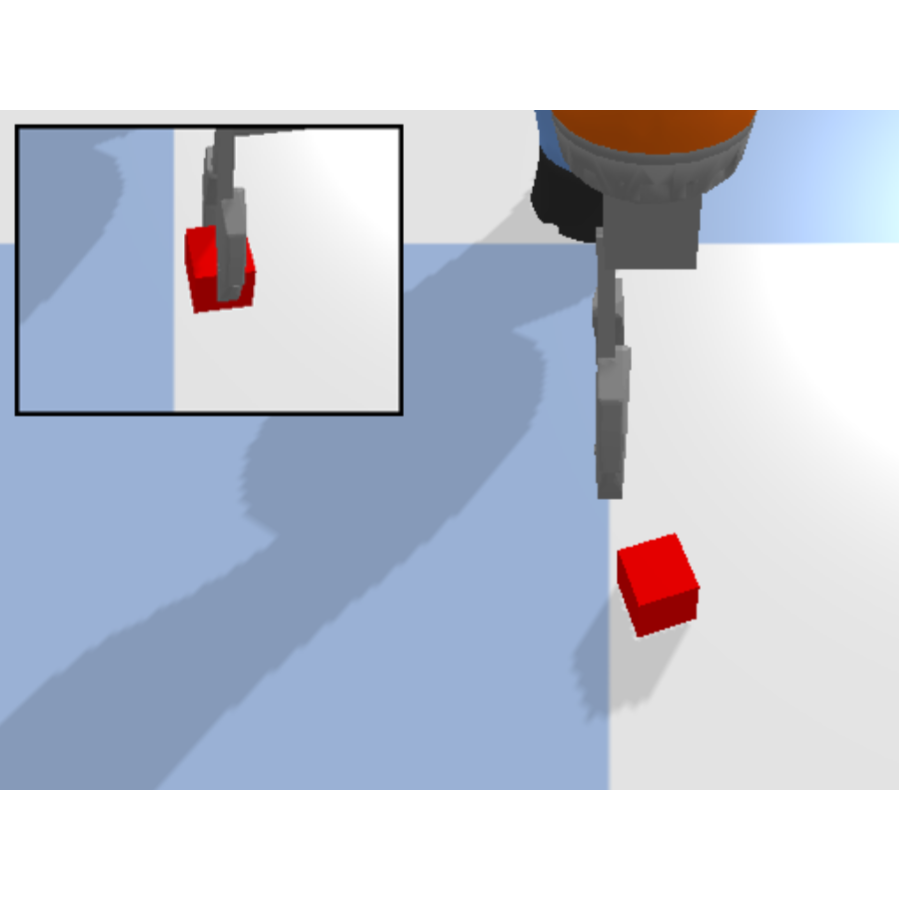}
    \caption{}
    \label{fig:sub6}
  \end{subfigure}
  \begin{subfigure}[b]{0.11\textwidth}
    \includegraphics[width=\linewidth,valign=t]{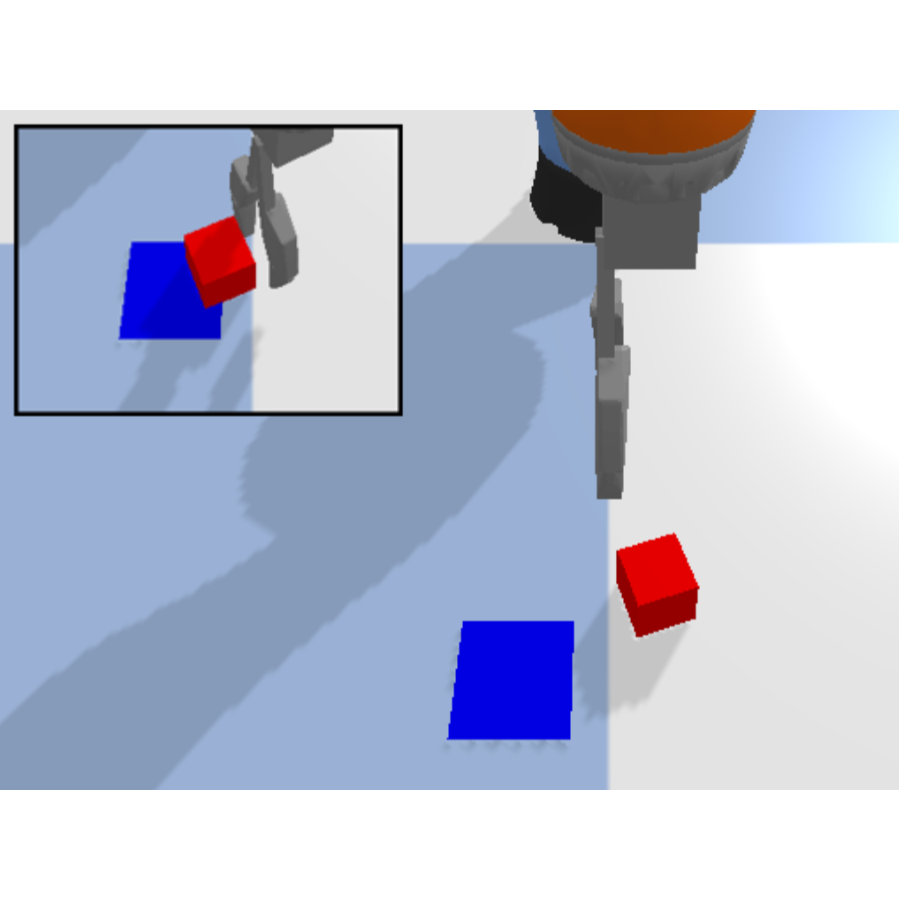}
    \caption{}
    \label{fig:sub7}
  \end{subfigure}
  \begin{subfigure}[b]{0.11\textwidth}
    \includegraphics[width=\linewidth,valign=t]{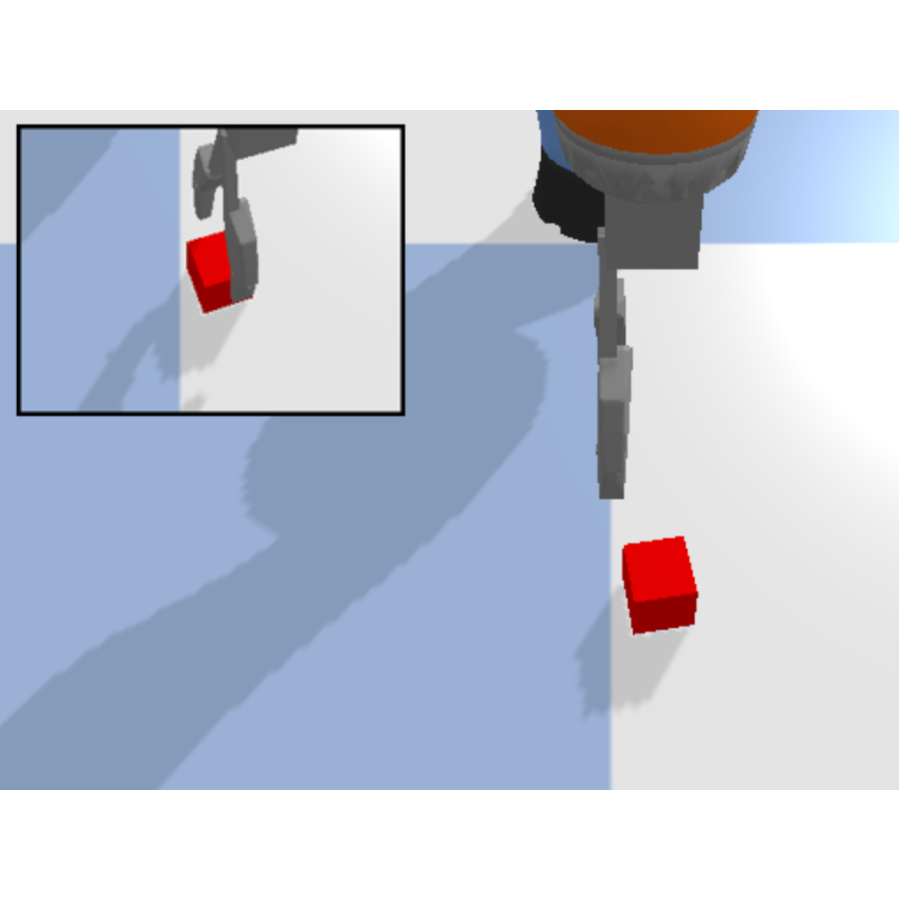}
    \caption{}
    \label{fig:sub8}
  \end{subfigure}
  \caption{Simulation environments: (a) Sawyer Bin, (b) Sawyer Push, (c) Fetch Pick And Place, (d) Fetch Push, (e) Fetch Reach, (f) BulletArm Block Pick, (g) BulletArm Block Push, (h) BulletArm Block Reach}
  \label{fig:three_images}
  \vspace{-15pt}
\end{figure}

\textbf{State Based Environments.}
\label{sec:state}
To ensure a fair comparison, we adopt the state-based setting from \citet{eysenbach_contrastive_2022}. As shown in Figure~\ref{fig:three_images}, we evaluate our methods and baselines on FetchReach and FetchPush from the MuJoCo-based Fetch suite~\citep{todorov_mujoco_2012}, and SawyerPush and SawyerBin from Meta-World~\citep{yu_meta-world_2019}. In addition, we introduce FetchPickAndPlace from the Fetch suite and BlockPush and BlockPick from the BulletArm~\citep{wang_bulletarm_2022}, which are not included in the original evaluation of \citep{eysenbach_contrastive_2022}.



\begin{figure}[t]
\vspace{-8pt}
  \centering


  \includegraphics[width=0.9\textwidth]{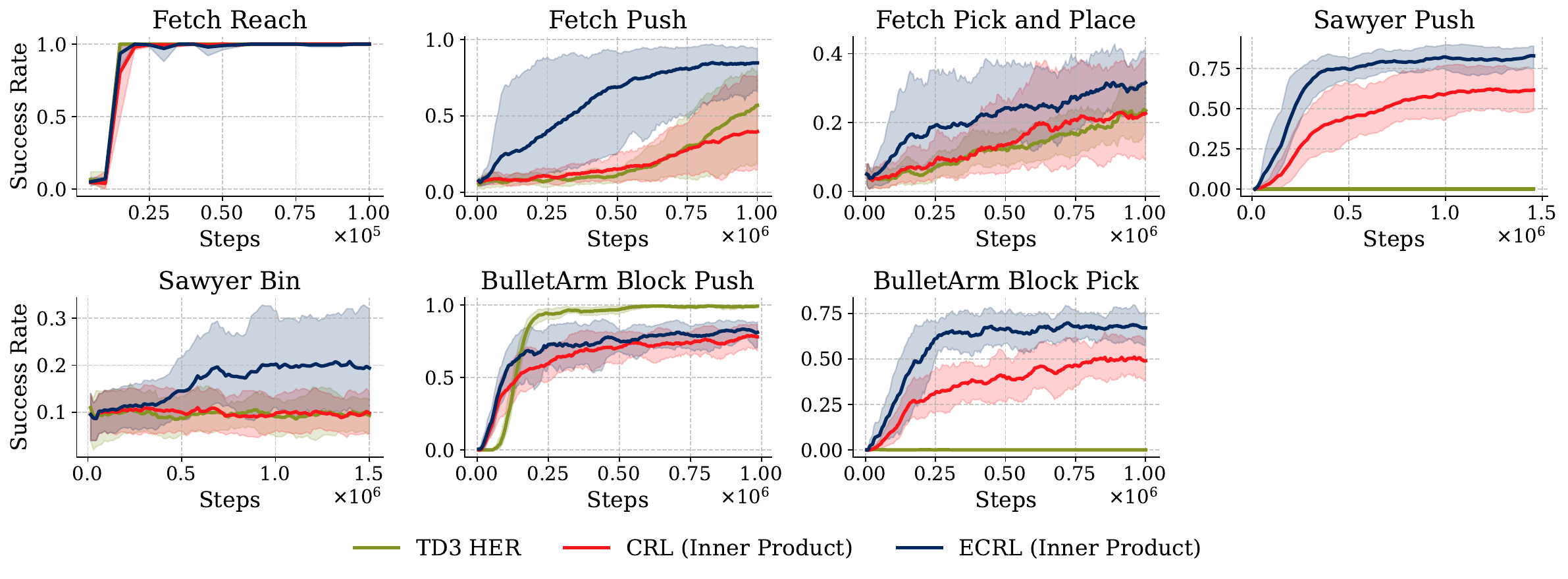}
    \caption{Comparison of Equivariant and Non-Equivariant Goal-Conditioned Contrastive RL on state-based tasks from the Fetch suite, Meta-World, and BulletArm. Results are averaged over \textbf{4 random seeds} for Meta-World and Fetch tasks, and \textbf{3 seeds} for BulletArm tasks.}

    \label{fig:main_state_tasks}
  \label{fig:main_state}
  \vspace{-8pt}
\end{figure}

Figure~\ref{fig:main_state} presents results on state-based tasks comparing our proposed method with CRL~\citep{eysenbach_contrastive_2022} and TD3+HER~\citep{andrychowicz_hindsight_2017}. We use inner-product as a similarity metric for these experiments. Our method consistently achieves higher final performance and greater sample efficiency across all tasks. While TD3+HER performs comparably to the non-equivariant variant in some environments, both are clearly outperformed by our equivariant approach, demonstrating the value of incorporating rotation-equivariance into goal-conditioned learning. A notable exception is BlockPick, a particularly challenging task where none of the methods succeed from scratch; to enable learning, we provide five expert demonstrations and include a behavioral cloning loss on the actor.

\begin{figure}[t]
  \centering


  \includegraphics[width=0.9\textwidth]{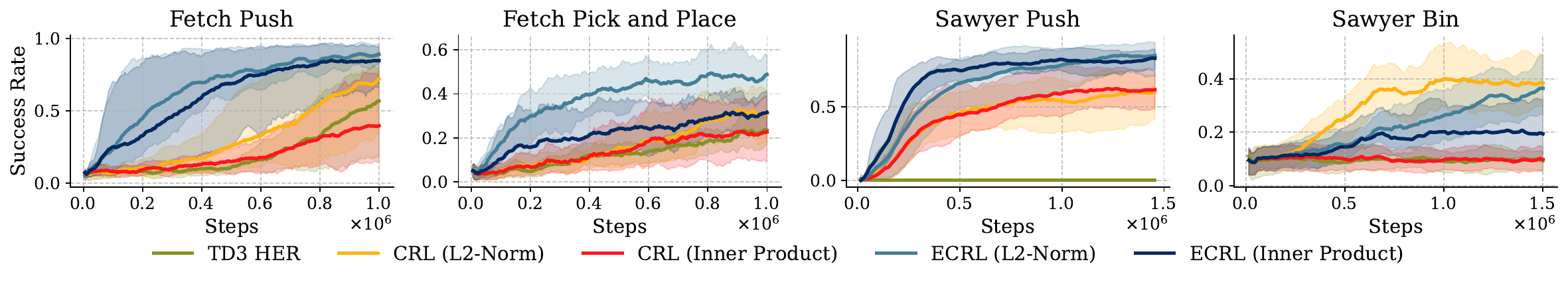}
  \caption{Comparison of Equivariant and Non-Equivariant Goal-Conditioned Contrastive RL on state-based pushing tasks, with and without L2-Norm as similarity metric. Results are averaged over \textbf{4 random seeds.}}

  \label{fig:l2_norm}
\vspace{-15pt}
\end{figure}

Following the $\ell_2$ metric used by \citet{bortkiewicz_accelerating_2024}, we compare two similarity metrics for the critic—inner product and $\ell_2$ distance—on state-based tasks from the Fetch and Meta-World benchmarks. As shown in Figure~\ref{fig:l2_norm}, both perform well when paired with our equivariant architecture, often outperforming the non-equivariant baseline, with $\ell_2$ distance slightly outperforming inner product on average. While both equivariant variants perform well overall, we observe that in SawyerBin, CRL (non-equivariant) with $\ell_2$ similarity achieves the best performance—likely due to limited variation in goal position, which reduces the benefit of rotation-equivariance.

\begin{figure}[htbp]
  \centering

  \begin{subfigure}[b]{0.9\textwidth}
    \includegraphics[width=\textwidth]{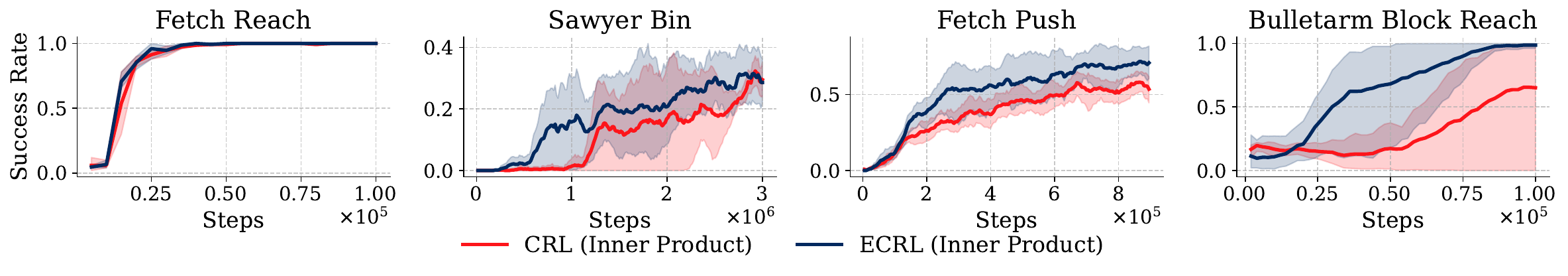}
  \end{subfigure}

  \caption{Equivariant contrastive learning consistently outperforms standard contrastive learning on image based tasks. Results are averaged over \textbf{3 random seeds.}}
  \label{fig:main_img}
\vspace{-10pt}
\end{figure}

\textbf{Image Based Environments.} For our image-based experiments, we evaluate on FetchReach and FetchPush from the Fetch suite, BlockReach from BulletArm, and SawyerBin from Meta-World. Our results are reported in Figure~\ref{fig:main_img}. Consistent with the state-based setting, we find that incorporating rotation-equivariant architectures in the image-based setting yields substantial gains in both final performance and sample efficiency. 
It is worth noting that the camera in these tasks is slightly skewed rather than strictly top-down, meaning the symmetry encoded in the network is only partially manifested in the image observations. Yet, ECRL learns more efficiently and achieves higher success rates than the baseline. These findings further align with prior studies \cite{wang_surprising_2022, wang_equivariant_2024} showing that equivariant methods often outperform non-equivariant ones, even when the encoded symmetry is only partially present, as is common in most real-world settings.

\vspace{-2pt}
\textbf{Rotation-Invariant Critic Representation.} While we achieve equivariance through aligned permutations of $\phi(s,a)$ and $\psi(\mathbf{g})$, a commonly used alternative approach is to use rotation-invariant networks for both, ensuring that their outputs remain unchanged under input rotations. This is typically done via global pooling over the group dimension of the regular representation features.

\begin{figure}[t]
  \centering
  \begin{subfigure}[b]{0.9\textwidth}
    \includegraphics[width=\textwidth]{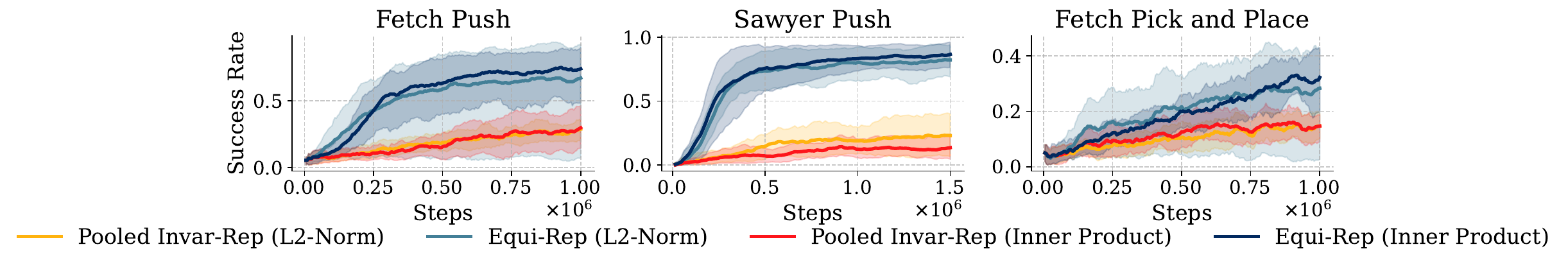}
  \end{subfigure}

  \caption{Achieving invariance through equivariant representations has much higher data efficiency than through global pooling. Results are averaged over \textbf{4 random seeds.}}
  \label{fig:pool}
  \vspace{-10pt}
\end{figure}

In this ablation study, we compare our structured equivariant critic to a pooled-invariant variant, where regular representation vectors are collapsed to trivial ones (1-element) through group-pooling, resulting in: : $\phi(s, a) \in \mathbb{R}^{1 \cdot K}$ and $\psi(\mathbf{g}) \in \mathbb{R}^{1 \cdot K}$. For fairness, both critics are paired with a standard, non-equivariant actor, isolating the effect of the critic’s representation of invariance.
As shown in Figure~\ref{fig:pool}, our equivariant critic substantially outperforms the pooled-invariant variant across both Fetch and Meta-World tasks. These results highlight that global pooling aggressively discards critical orientation information, severely limiting the quality of learned representations for Contrastive RL, and underscore the importance of preserving equivariant structure for downstream performance.


\setlength{\tabcolsep}{4pt} 
\begin{wraptable}{r}{0.55\textwidth}
    \centering
    \scriptsize
    \renewcommand{\arraystretch}{1.5}
    \begin{tabular}{ll|cc}
        \toprule
        \textbf{Task} & \textbf{Dataset Size} & \textbf{ECRL (Ours)} & \textbf{CRL} \\
        \midrule
        \multirow{2}{*}{FetchReach} 
            & 5 demos  & \textbf{0.90} $\pm$ 0.06 & 0.57 $\pm$ 0.12 \\ 
            & 10 demos & \textbf{0.97} $\pm$ 0.02 & 0.90 $\pm$ 0.05 \\
        \midrule
        \multirow{2}{*}{FetchPush} 
            & 10 demos & \textbf{0.36} $\pm$ 0.03 & 0.28 $\pm$ 0.01 \\
            & 50 demos & \textbf{0.47} $\pm$ 0.02 & 0.41 $\pm$ 0.03 \\
        \midrule
        \multirow{2}{*}{SawyerPush} 
            & 10 demos & \textbf{0.60} $\pm$ 0.04 & 0.45 $\pm$ 0.05 \\
            & 50 demos & \textbf{0.73} $\pm$ 0.04 & 0.68 $\pm$ 0.02 \\
        \bottomrule
    \end{tabular}
    \caption{Performance comparison between ECRL and CRL across different dataset sizes. Bold indicates the best performance in each row. Results are averaged over \textbf{4 random seeds.}}
    \label{tab:ecrl_vs_crl}
\end{wraptable}
\textbf{Offline Contrasive RL.} 
To validate our method in offline RL settings, 
we first collect sub-optimal datasets using policies trained with Equivariant Contrastive RL. For FetchPush and SawyerPush, we evaluate performance with datasets of 10 and 50 demonstrations, while for FetchReach, we use smaller datasets of 5 and 10 demonstrations due to the task's simplicity. 
Following \citet{eysenbach_contrastive_2022} and \citet{fujimoto2021minimalist}, we adapt Contrastive RL to the offline setting by adding a goal-conditioned behavioral cloning term to the policy objective (Equation~\ref{Offline_RL_loss}), and adopt their recommended hyperparameters. We use inner product as the similarity metric and report performance using success rate as the evaluation metric. Policies are evaluated every 1000 gradient updates, and we report the success rate of the best-performing checkpoint, averaged over 4 random seeds. For each evaluation, we run 100 episodes and report the average success rate.
As shown in Table~\ref{tab:ecrl_vs_crl}, ECRL consistently outperforms the non-equivariant baseline across all tasks and dataset sizes, with particularly strong gains in low-data regimes.

\normalsize

%% file: conclusion.tex
\vspace{-8pt}
In this work, we introduce the Goal-Conditioned Group-Invariant MDP and propose Equivariant Contrastive RL (ECRL), a framework that leverages rotational symmetries in robotic manipulation tasks to improve sample efficiency and generalization. At its core is a novel rotation-invariant critic that enables effective symmetry exploitation within the contrastive learning framework. Our experiments demonstrate that ECRL consistently outperforms standard Contrastive RL across both state- and image-based settings, and that our equivariant representation-based approach significantly outperforms alternatives such as global pooling.

While ECRL excels on tasks with rotational symmetry, its benefits diminish in settings with minimal or no goal variation under symmetry, as seen in SawyerBin. Extending ECRL to handle richer symmetry groups, such as $SE(3)$, could further improve generalization across diverse tasks. Additionally, our method for enforcing rotation-invariance may benefit other representation learning approaches, such as \citep{park_metra_2024, ma2022vip, park2024foundation, nair2022r3m}, which rely on element-wise operations.

%% file: appendix.tex
\section{Proof}

First, we define a new MDP, $\hat{\mathcal{M}}$, with state space $\hat{S} := S \times \mathbf{G}$. The group action on this state space is $g \cdot (s, \mathbf{g}) = (gs, g\mathbf{g})$, and the group action on the action space is $g \cdot a$. 

The transition function of this augmented MDP is equivariant by definition (Section \ref{sec:GC-G-invariant MDP}): 
\begin{equation}
    p(\hat{s}' | \hat{s}, a)
 = p((s', \mathbf{g}) | (s, \mathbf{g}), a) = p((gs', g\mathbf{g}) | (gs, g\mathbf{g}), ga) = p(g\hat{s}' | g \hat{s}, ga)
\end{equation}

and the reward function is invariant,

\begin{equation}
    r(\hat{s}, a) = r((s, \mathbf{g}), a) = r((gs, g\mathbf{g}), ga) = r(g\hat{s}, ga).
\end{equation}

Applying Proposition 4.1 from~\citep{wang_mathrmso2-equivariant_2022}, we know that the optimal $Q$-function and policy for $\hat{\mathcal{M}}$ are invariant and equivariant respectively. That is,

\begin{equation}
    \hat{Q}^*(g\hat{s}, ga) = \hat{Q}^*(\hat{s}, a),
\end{equation}

is invariant and 

\begin{equation}
    \hat{\pi}^*(g\hat{s}) = g\hat{\pi}^*(\hat{s})
\end{equation}

is equivariant.

Thus, applying the group actions to the elements of our augmented state space we can conclude that the goal-conditioned policy and $Q$-function are also invariant, 

\begin{equation}
    Q^*(gs, ga, g\mathbf{g}) = Q^*(s, a, \mathbf{g})
\end{equation}

and equivariant

\begin{equation}
    \pi^*(gs, g\mathbf{g}) = g\pi^*(s, \mathbf{g}).
\end{equation}

Thus, the optimal policy and $Q$-function of a goal-conditioned group-equivariant MDP are equivariant and invariant respectively.

\section{Implementation and Experiment Details}

\subsection{Network Architecture and Hyperparameters}
We implement the networks for Contrastive RL and all other baselines in PyTorch. For Equivariant Contrastive RL, we use the \texttt{e2cnn} library to construct $C_N$-equivariant neural networks. Unless otherwise specified, we adopt the same hyperparameters for all algorithms as reported in \citet{eysenbach_contrastive_2022}. Additionally, we use the same network architectures for Contrastive RL and other baselines as in the original implementation by \citet{eysenbach_contrastive_2022}.


For Equivariant Contrastive RL, the state-action and the goal encoders are implemented as $C_8$-equivariant MLPs for the state-based tasks. Each encoder takes as input a mixed-representation $1 \times 1$ feature map, consisting of $\rho_1$ features for the equivariant components and $\rho_0$ features for the invariant components of the state or goal. The equivariant MLPs have two hidden layers, each outputting 256$\times$8-dimensional regular representation features.  The final layer outputs a 64$\times$8-dimensional regular representation embedding (i.e., a 64-channel $1 \times 1$ feature map with 8-dimensional regular representations). For the similarity metric computation, each corresponding element across the $(s,a)$ and $\mathbf{g}$ regular representation embeddings is compared—using elementwise products for inner product similarity or squared differences for $\ell_2$ distance. For example, the third element of the second regular representation in $\phi(s,a)$ interacts with the third element of the second regular representation in $\psi(\mathbf{g})$. 

The policy network follows a similar equivariant architecture: it takes as input a mixed-representation $1 \times 1$ feature map constructed from the state and goal, processes it through two hidden layers of 256 regular representation features, and outputs a mixed-representation $1 \times 1$ feature map corresponding to the action, containing both $\rho_1$ and $\rho_0$ components.

For image-based tasks, we use a $C_8$-equivariant steerable CNN as the image encoder. 
Each of the actor and critic networks uses a dedicated equivariant image encoder shared across observations and goals. These encoders take each observation or goal image as a 3-channel $\rho_0$ input and produce a stack of 64 $1 \times 1$ regular representation feature embeddings. These embeddings are then used in place of raw state or goal vectors and passed into two-layer $C_8$-equivariant MLPs, mirroring the architecture described above for the state-based encoders and policy networks.

\begin{table}[htbp]
    \centering
    \small
    \renewcommand{\arraystretch}{1}
    \setlength{\tabcolsep}{8pt}
    \begin{tabular}{l|l}
        \toprule
        \textbf{Hyperparameter} & \textbf{Value} \\
        \midrule
        batch size & 256 for state-based, 64 for image-based\\
        learning rate & 3e-4 for all components \\
        discount & 0.99 \\
        actor target entropy & 0 for state-based, $-\mathrm{dim}(a)$ for image-based \\
        target EMA term (for TD3 and SAC) & 0.005 \\
        hidden layer sizes (actor and critic networks) & (256, 256) \\
        initial random data collection & 10,000 transitions \\
        replay buffer size & 1,000,000 transitions \\
        train-collect interval & 16 for state-based, 64 for image-based \\
        representation dim (dim$(\phi(s,a)), \dim(\psi(s_g))$) & 64 \\
        actor minimum std dev & 1e-6 \\
        goals for actor loss & random states (not future states) \\
        equivariant rotation group & $C_8$ \\
        \bottomrule
    \end{tabular}
    \vspace{5pt}
    \caption{Hyperparameters for our method and the baselines.}
    \label{tab:hyperparams}
    \vspace{-10pt}
\end{table}

\begin{table}[htbp]
    \centering
    \small
    \renewcommand{\arraystretch}{1}
    \begin{tabular}{l|l}
        \toprule
        \textbf{hyperparameter} & \textbf{value} \\
        \midrule
        batch size & 256 $\rightarrow$ 1024 for CRL, 256 $\rightarrow$ 512 for ECRL\\
        representation dimension & 64 $\rightarrow$ 16 \\
        hidden layers sizes (for actor and representations) & (256, 256) $\rightarrow$ (1024, 1024) for CRL, \\ & (256, 256) $\rightarrow$ (256, 256) for ECRL \\
        goals for actor loss & future states \\
        \bottomrule
    \end{tabular}
    \vspace{5pt}
    \caption{Modified hyperparameters in the offline experiments.}
    \label{tab:offline_hyperparams}
\end{table}

\subsection{Implementation Details}
\label{appendix_sub_sec:Implementation Details}
We adopt the same implementation details for Contrastive RL and other baselines from \citet{eysenbach_contrastive_2022}, and highlight the key modifications made for our experiments below.

\textbf{Similarity Metric for CRL}: As mentioned in Sec.~\ref{sec:method}, we explore two similarity metrics in our work: inner product ($f(s, a, \mathbf{g}) := \phi(s, a)^\intercal \psi(\mathbf{g})$) and $\ell_2$ distance ($f(s, a, \mathbf{g}) := -\|\phi(s, a) - \psi(\mathbf{g})\|_2$). Following the approach of \citet{park2022probabilistic}, we define the $\ell_2$-based similarity metric as the negative squared distance between the embeddings, scaled and shifted by learnable parameters:
\[
f(s, a, \mathbf{g}) = -a \cdot \|\phi(s, a) - \psi(\mathbf{g})\|_2 + b
\]
where $a$ and $b$ are scalar parameters jointly optimized with the critic. We found this form to yield more stable learning than using a raw $\ell_2$ distance, while preserving the contrastive structure of the method.

\textbf{HER Relabeling Strategy:} To better highlight the benefits of incorporating equivariance into CRL, our main results (Sec.~\ref{sec:experiments}) use a strong baseline setup in which TD3+HER is granted access to the ground-truth reward function for relabeling—unlike prior work such as \citet{eysenbach_contrastive_2022, wang2023optimal}. We find that, although TD3+HER performs competitively against CRL under this privileged setting, ECRL consistently outperforms both baselines.

\subsection{Environments}
In this subsection, we describe the environments used in our evaluation that were not included in \citet{eysenbach_contrastive_2022}:

\textbf{fetch pick and place} (state): This task requires the robotic arm to pick up an object and place it at a specified goal location in free space. A rollout is considered successful if the object is placed within 5cm of the goal position.

\textbf{bulletarm block push} (state): Similar to the fetch push task from the Fetch suite and sawyer push from Meta-World, this task from \citet{wang_bulletarm_2022} involves using a robotic arm to push a block to a specified goal location on the table. Success is defined as moving the object within 5cm of the goal.

\textbf{bulletarm block pick} (state): Similar to the fetch pick and place task from the Fetch suite, this task from \citet{wang_bulletarm_2022} involves using a robotic arm to pick up a block from one location and place it at a specified goal position. The goal location varies in the XY plane but remains fixed in height (Z-axis). An episode is considered successful if the block is placed within 5 cm of the goal position.

\textbf{bulletarm block reach} (image): Similar to the fetch reach task from the Fetch suite, this task from \citet{wang_bulletarm_2022} involves guiding the robot’s end-effector to a target location marked by a block in the workspace. While the task is straightforward, it poses a unique challenge due to the use of gripper-centered top-down depth images for both observations and goals—resulting in visually similar goal images where the block is always centered beneath the gripper in the center.

\section{Additional Baseline Experiments with TD3 + HER}
As discussed in Appendix~\ref{appendix_sub_sec:Implementation Details}, the relabeling strategy used by TD3+HER in our state-based experiments assumes access to the ground-truth reward function. However, this is not feasible in settings with image-based observations, where computing reward based on pixel-level similarity is unreliable. To evaluate TD3+HER in a more realistic setting, we conduct experiments under a constrained relabeling strategy that only assigns positive reward when the achieved state exactly matches the goal state. We report results on three tasks—FetchReach, FetchPush, and SawyerBin—to assess the robustness of TD3+HER when explicit reward supervision is unavailable. We further evaluate TD3+HER with this restricted relabeling strategy on two state-based tasks—FetchPush and FetchPickAndPlace — where it performs competitively with standard CRL.

\begin{figure}[t]
  \centering
  \begin{subfigure}[b]{0.9\textwidth}
    \includegraphics[width=\textwidth]{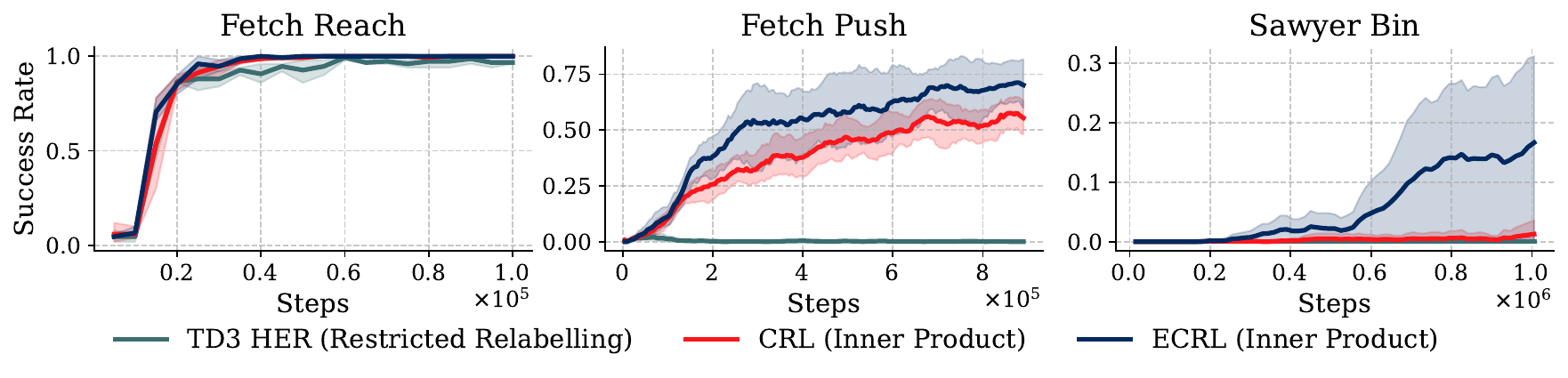}
  \end{subfigure}

  \caption{Comparison of TD3+HER, CRL, and ECRL on Image-Based Tasks. Both CRL and ECRL outperform TD3+HER, with ECRL achieving the best overall performance and consistency. Results are averaged over \textbf{3 random seeds.}}

  \label{fig:new_td3_her_image}
  \vspace{-10pt}
\end{figure}

As shown in Figure~\ref{fig:new_td3_her_image}, TD3+HER is able to learn a reasonable policy on the FetchReach image-based task, but fails to make meaningful progress on the other evaluated tasks. Both CRL and ECRL outperform TD3+HER, with ECRL demonstrating the most consistent and robust performance across all tasks.

\begin{figure}[htbp]
  \centering
  \begin{subfigure}[b]{0.9\textwidth}
    \includegraphics[width=\textwidth]{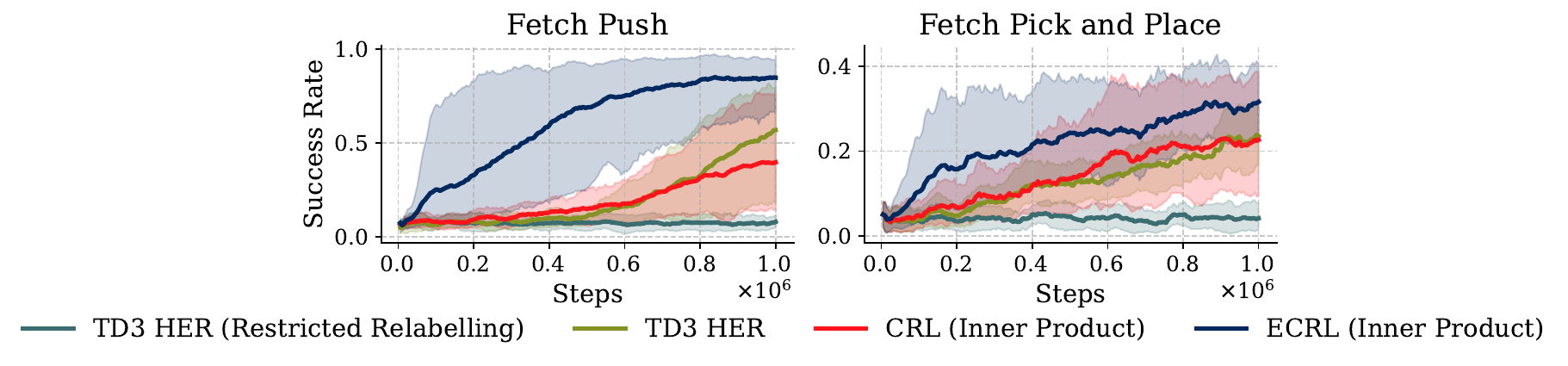}
  \end{subfigure}

  \caption{Comparison of TD3+HER with the restricted relabeling strategy on state-based tasks. The strategy leads to significantly lower returns compared to relabeling using the ground-truth reward function, with ECRL performing the best. Results are averaged over \textbf{3 random seeds.}}

  \label{fig:new_td3_her_state}
  \vspace{-10pt}
\end{figure}

For further evaluation of TD3+HER with this relabeling strategy on state-based tasks, we find that its performance degrades significantly compared to when it has access to the ground-truth reward function for relabeling, as can be seen in Figure~\ref{fig:new_td3_her_state}.

\section{Generalization to Unseen Goals} 
\begin{figure}[t]
  \centering
  \begin{subfigure}[b]{0.6\textwidth}
    \includegraphics[width=\textwidth]{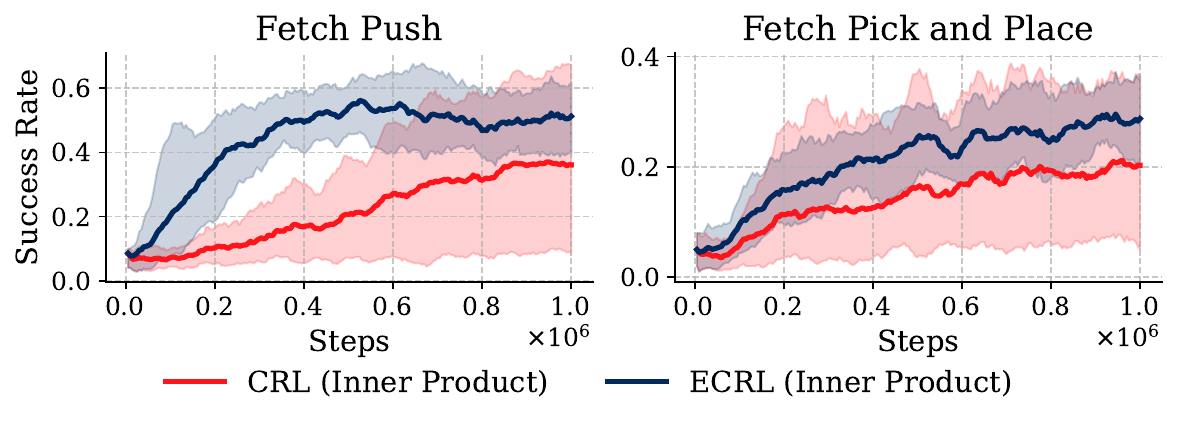}
  \end{subfigure}

  \caption{Generalization to unseen goals sampled outside the training quadrant. ECRL generalizes better and exhibits lower variance compared to CRL, highlighting the benefits of incorporating rotation-equivariance. Results are averaged over \textbf{3 random seeds.}}
  \label{fig:one_quadrant}
  \vspace{-10pt}
\end{figure}

In this section, we assess how the incorporation of rotation-equivariance impacts generalization to novel, unseen goals. During training, we restrict the goal positions to the first quadrant of the workspace by sampling the goal X and Y offsets uniformly from $[0, r]$, where $r$ is the environment's target range. At evaluation time, however, goals are sampled from the full workspace, i.e., $[-r, r]$, to test generalization to unseen goal regions. We report results on two state-based tasks—FetchPush and FetchPickAndPlace.

Figure~\ref{fig:one_quadrant} showcases our results. ECRL continues to outperform CRL, demonstrating better generalization to novel, unseen goals. We observe that both ECRL and CRL experience a drop in performance compared to the original setting where goals are sampled from the entire workspace during training (Figure~\ref{fig:main_state_tasks}). Additionally, we observe that ECRL exhibits greater stability, with significantly lower variance across 4 random seeds compared to CRL. These results provide strong evidence that incorporating rotation-equivariance improves both the robustness and generalization of goal-conditioned policies.





\section{{Additional Results with InfoNCE Loss}}
In this section, we present additional experiments using the InfoNCE loss as the contrastive objective, instead of the original Binary NCE loss formulation used in the main paper. This variant of has been widely used in several recent works that have used Contrastive-RL \cite{zheng2023contrastive, bortkiewicz_accelerating_2024, liu_single_2024}. We evaluate both CRL and our proposed ECRL under this formulation, keeping all other hyperparameters and architectural components unchanged.

\begin{figure}[h]
  \centering
  \begin{subfigure}[b]{0.9\textwidth}
    \includegraphics[width=\textwidth]{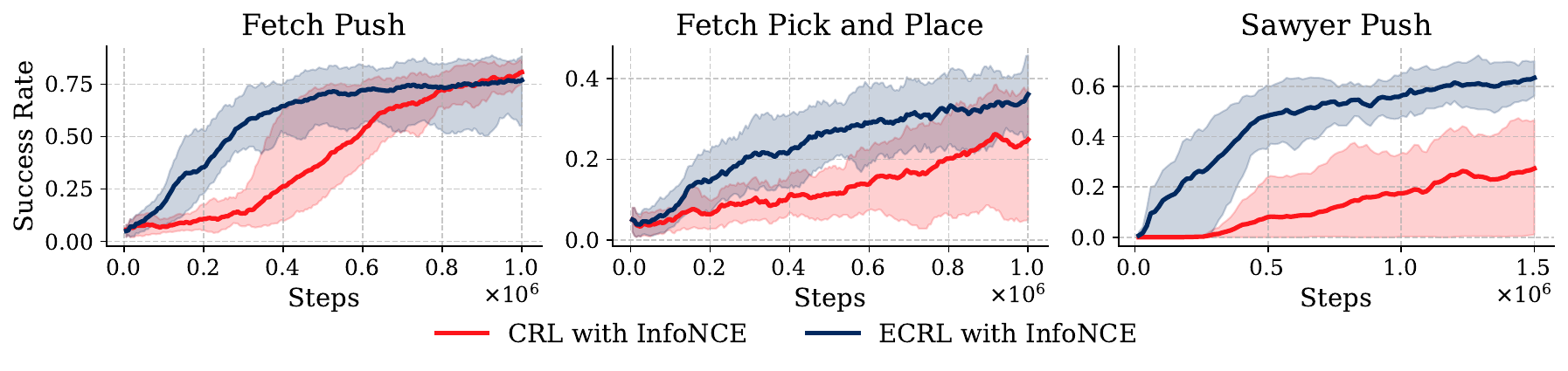}
  \end{subfigure}

  \caption{{Comparison of ECRL and CRL using the InfoNCE loss across various tasks. ECRL consistently outperforms CRL, demonstrating that the benefits of equivariance are orthogonal to the choice of contrastive loss. Results are averaged over \textbf{3 random seeds.}}}
  \label{fig:infonce_state}
  \vspace{8pt}
\end{figure}

As shown in Fig~\ref{fig:infonce_state}, ECRL continues to outperform CRL across all tasks, reaffirming that the performance gains stem from incorporating group-equivariant structure, and are orthogonal to the choice of contrastive loss.

\section{Additional Ablations}
To further understand the sensitivity of ECRL to the design of its equivariant components, we conduct ablations over (i) the order (N) of the cyclic group \(C_N\) and (ii) the number of regular representations \(K\) in the output representations of $(s,a)$ and $\mathbf{g}$.

\subsection{Number of Regular Representations (K)}

\begin{figure}[h]
  \centering
  \begin{subfigure}[b]{0.7\textwidth}
    \includegraphics[width=\textwidth]{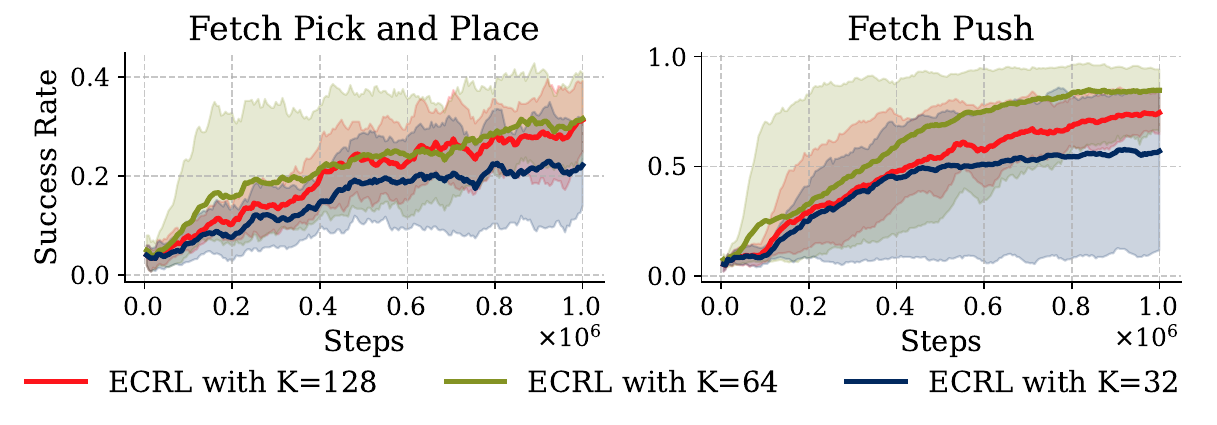}
  \end{subfigure}

  \caption{{Ablation on the number of regular representations \(K\) in ECRL. \(K=64\) offers the best balance between performance and computational efficiency, with \(K=32\) underfitting and \(K=128\) providing minimal gains. Results are averaged over \textbf{3 random seeds.}}}

  \label{fig:k_ablation}
\end{figure}

Fig~\ref{fig:k_ablation} shows the performance of ECRL with \(K \in \{32, 64, 128\}\), where \(K=64\) is used in the main paper. Increasing \(K\) to 128 provides only marginal improvements on the Fetch Push task and negligible gains on Fetch Pick and Place, while incurring a significant computational overhead. Conversely, the reduced performance with \(K=32\) suggests that insufficient representational capacity limits the encoders' ability to learn useful state-action and goal representations.

\subsection{Cyclic Group Order (N)}

\begin{figure}[h]
  \centering
  \begin{subfigure}[b]{0.7\textwidth}
    \includegraphics[width=\textwidth]{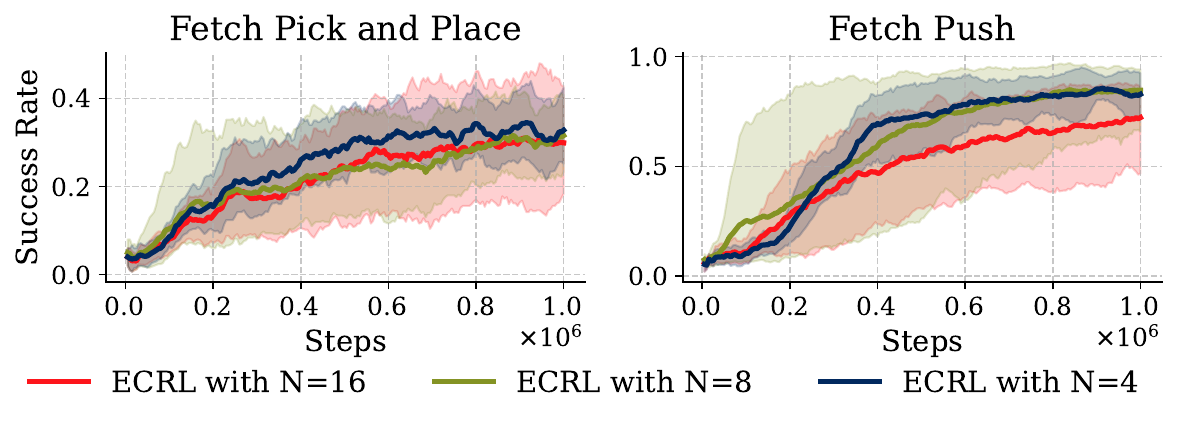}
  \end{subfigure}

  \caption{{Ablation on the cyclic group order \(N\) in ECRL. \(N=4\), \(N=8\), and \(N=16\) perform comparably, with \(N=8\) used in the main paper. Results are averaged over \textbf{3 random seeds.}}}
  \label{fig:n_ablation}
\end{figure}

Figure~\ref{fig:n_ablation} shows the performance of ECRL with \(N \in \{4, 8, 16\}\), where \(N=8\) is used in the main paper. The fact that \(N=4\) and \(N=16\)  performs comparably to \(N=8\) suggests that ECRL is robust to the choice of group order $N$ for these tasks. We do observe a slight drop in performance with \(N=16\) for the Fetch Push task, but this drop is statistically insignificant.

\section{Extensibility to Other Symmetry Groups}
In the main paper, we focus on planar rotational equivariance, since common tabletop manipulation tasks (e.g., block pushing) predominantly involve planar motion and naturally exhibit approximate $SO(2)$ symmetry. However, our theoretical results (Proposition~\ref{prop1}) are group-agnostic, and the proposed architectures readily extend to other discrete symmetry groups, such as the $D_4$ group (rotations and reflections) and discretized 3D rotations (e.g., the octahedral subgroup of $SO(3)$). Extending to these settings requires only replacing the group definition in the equivariant networks\cite{cesa_program_2022,weiler_general_2019}, with no change to the overall training pipeline.

To demonstrate the extensibility of our framework, we apply $D_4$-ECRL (rotations and reflections) on the Sawyer Push task and Octahedral-ECRL (leveraging the octahedral subgroup (24-discretizations) of $SO(3)$) on the Fetch Reach task, showing that our method can seamlessly handle richer symmetry groups without any modifications to the training pipeline.

\begin{figure}[h]
  \centering
  \begin{subfigure}[b]{0.9\textwidth}
    \includegraphics[width=\textwidth]{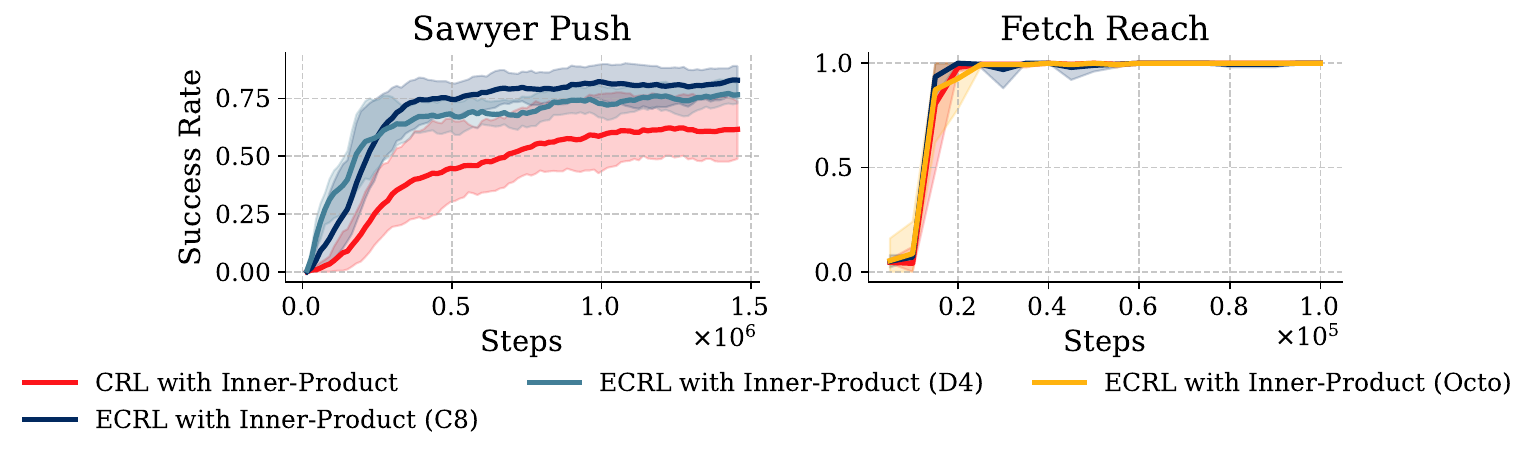}
  \end{subfigure}

  \caption{{Demonstration of ECRL with different symmetry groups. We show results for $D_4$-ECRL on Sawyer Push and Octahedral-ECRL on Fetch Reach, highlighting that our framework extends seamlessly beyond $C_8$. Results are averaged over \textbf{3 random seeds.}}}
  \label{fig:different_symmetry_groups}
\end{figure}

Fig.~\ref{fig:different_symmetry_groups} showcases that our approach seamlessly adapts to different discrete symmetry groups, with $D_4$-ECRL effectively exploiting the reflectional and rotational symmetries in Sawyer Push and outperforming non-equivariant CRL. However, we observe no significant performance gain over $C_8$-ECRL for Sawyer Push. For Fetch Reach, we use the octahedral subgroup of $SO(3)$, which consists of 24 rotation elements. Similar to $C_8$-ECRL and CRL, Octo-ECRL also achieves optimal performance. We restrict the evaluation to Fetch Reach due to the significantly increased training time from the added computational overhead, as well as the relatively short time required to achieve optimal performance on this task.